\documentclass[conference]{IEEEtran}
\IEEEoverridecommandlockouts
\usepackage{cite}
\usepackage{amsmath,amssymb,amsfonts}
\usepackage{algorithmic}
\usepackage{graphicx}
\usepackage{textcomp}
\usepackage{xcolor}
\usepackage{multirow}
\usepackage{url}
\def\BibTeX{{\rm B\kern-.05em{\sc i\kern-.025em b}\kern-.08em
    T\kern-.1667em\lower.7ex\hbox{E}\kern-.125emX}}
\begin{document}

\title{EMMeTT: Efficient Multimodal Machine Translation Training}

\author{
    \IEEEauthorblockN{Piotr Żelasko\textsuperscript{1},  Zhehuai Chen\textsuperscript{1}, Mengru Wang, Daniel Galvez, Oleksii Hrinchuk,  \\ Shuoyang Ding, Ke Hu, Jagadeesh Balam, Vitaly Lavrukhin, Boris Ginsburg}
    \IEEEauthorblockA{
        NVIDIA \\
        \{pzelasko,zhehuaic\}@nvidia.com
    }
}

\maketitle

\stepcounter{footnote}
\footnotetext{These authors contributed equally to this work.}

\begin{abstract}
A rising interest in the modality extension of foundation language models warrants discussion on the most effective, and efficient, multimodal training approach. 
This work focuses on neural machine translation (NMT) and proposes a joint multimodal training regime of Speech-LLM to include automatic speech translation (AST).
We investigate two different foundation model architectures, decoder-only GPT and encoder-decoder T5, extended with Canary-1B's speech encoder.
To handle joint multimodal training, we propose a novel training framework called EMMeTT.
EMMeTT improves training efficiency with the following: balanced sampling across languages, datasets, and modalities; efficient sequential data iteration; and a novel 2D bucketing scheme for multimodal data, complemented by a batch size optimizer (OOMptimizer).
We show that a multimodal training consistently helps with both architectures. Moreover, SALM-T5 trained with EMMeTT retains the original NMT capability while outperforming AST baselines on four-language subsets of FLORES and FLEURS. The resultant {\em Multimodal Translation Model} produces strong text and speech translation results at the same time.
\end{abstract}


\section{Introduction}

The ubiquitous availability and impressive capabilities of large language models (LLM) motivate our community to extend them to other modalities, and in the context of this work, audio modality in particular. 
There have been several recent works proposing different approaches to training a Speech-LLM: %
Speech-LLaMA and variants~\cite{chen2024salm,tang2023salmonn,qwen2023} directly concatenate the speech prompts with text prompts and use the combined sequence as input to the LLM like a decoder-only GPT model. Flamingo and its extension~\cite{kong2024audio,alayrac2022flamingo,chen2024bestow} are another branch of works, where 
cross-modal cross-attention is added to the pretrained GPT-based LLM with shared text query.

The central focus of this work is the study of multimodal training approach itself.
Firstly, the extension to audio modality typically causes catastrophic forgetting of the text domain capabilities. \cite{tang2023salmonn} analyzes the task over-fitting problem in Speech-LLM and proposes an activation tuning stage to alleviate this problem.
The task over-fitting limits the usefulness of the original LLM exclusively to the tasks seen in audio domain training, which may be viewed as limiting of the model's generalization, zero-shot, and in-context learning capabilities.

One possible solution to catastrophic forgetting is to train the model on data from multiple modalities jointly. 
Hence, the second point of interest is how to perform joint multimodal training effectively and efficiently.
Since the scope of such problem is very broad, we focus on a well-defined domain of neural machine translation (NMT) and demonstrate a solution that allows a model to retain text NMT capabilities while acquiring automatic speech translation (AST) skills.

Our main contribution is a joint multimodal training framework named EMMeTT that allowed us to train an NMT model with AST and NMT data jointly, resulting in a single model capable of performing both NMT and AST.
The resulting model matches the single-modal baseline NMT performance while outperforming AST baselines on English, French, German, and Spanish subset of FLEURS. 
To enable this, we design data sampling strategies from the first principles to ensure a stationary data distribution across modalities, languages, and datasets throughout the training.
Furthermore, we stratify the sampling with dynamic bucketing and a novel 2D bucketing technique for maximal training efficiency, complemented by a batch size optimizer (OOMptimizer) algorithm that pushes GPU memory utilization to its limits across various sequence lengths, while preventing spurious out-of-memory (OOM) errors. 
Finally, we demonstrate EMMeTT's flexibility by training two different multimodal model architectures: BESTOW-GPT and SALM-T5 and producing strong results.

\section{Related work}

\textbf{Speech and text translation.} 
Recent works leveraging NMT data to complement AST systems can be broken down into two paradigms: pretrain-finetune and joint training. 
Most works fall under pretrain-finetune category, e.g. \cite{tang2022unified} jointly pretrains speech and text in an encoder-decoder model for ASR and AST with self-supervised and supervised losses. 
\cite{zheng2021fused}  jointly learns a unified speech and text representation, which is later used to initialize an AST system.
Recent LLM-based systems usually connect a pretrained speech encoder with an NMT pretrained LLM. After finetuning with AST data, the systems can leverage speech and text pretrained knowledge from the encoder and decoder respectively and have shown strong performances, e.g. \cite{huang2023speech,xu2024cmu}. 
As for joint training, \cite{zhang2023google} conducted NMT and AST joint finetuning on the pretrained encoder and decoder with a learned upsampling text encoder. 
An ablation study in~\cite{bapna2022mslam} clearly shows the benefit of joint training in the last stage compared to speech-only finetune.

\textbf{Foundation models for translation.} Large foundation models have been successfully adapted for NMT tasks across diverse language pairs. Raffel et al.~\cite{raffel2020T5} introduced the T5 architecture, demonstrating its versatility in various text-to-text applications, including NMT. Brown et al. \cite{gpt3} presented GPT-3, a decoder-only LLM capable of few-shot learning in translation tasks, showcasing the potential of general-purpose language models in NMT. Advancing the field further, Liu et al.~\cite{liu2020mbart} extended BART~\cite{lewis2020bart} to multilingual settings, showing its capacity to handle translation across multiple languages with minimal fine-tuning. As the scope of NMT expanded, researchers began exploring multimodal extensions of these models, integrating additional modalities such as speech. Prior works, such as Speech Augmented Language Models (SALM)~\cite{chen2024salm} and BESTOW~\cite{chen2024bestow} demonstrated the potential for cross-modal learning, where text pretrained LLMs could be adapted to transfer knowledge acquired from massive text-based language resources to other modalities.

\section{Methods}

\begin{figure*}[!tbh]
    \centering
    \includegraphics[width=0.99\linewidth]{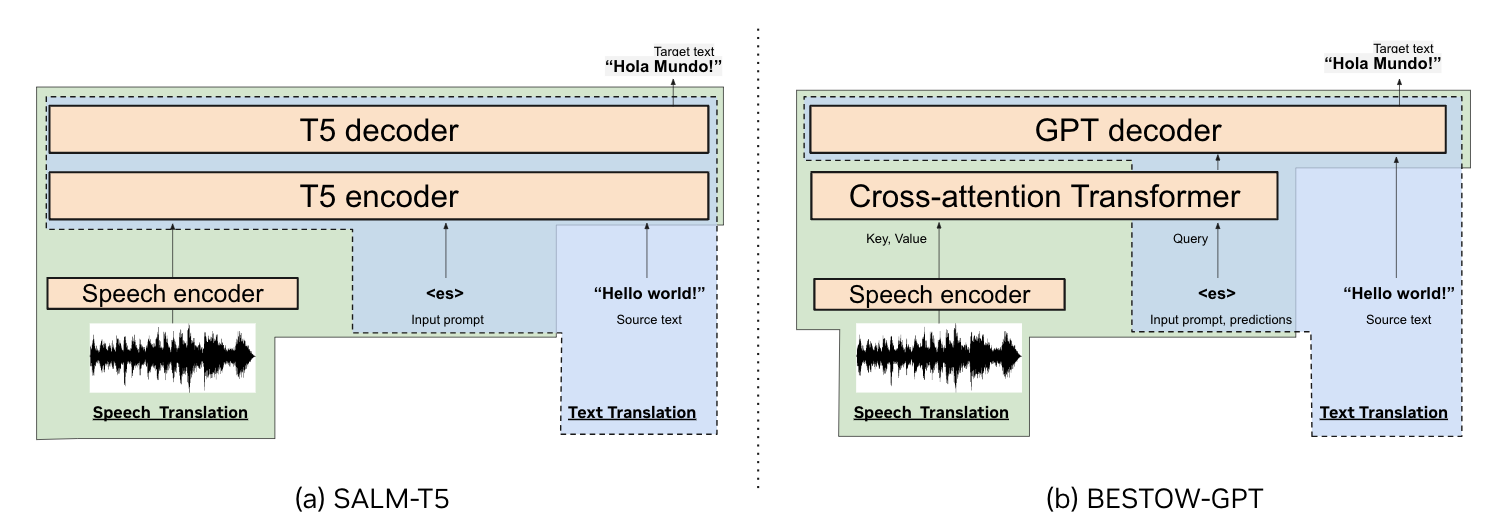}
    \vspace{-1em}
    \caption{(a) SALM-T5 architecture. (b) BESTOW-GPT architecture. The proposed multimodal joint training is visualized on both architectures where green shade denotes speech translation training and blue dotted shade denotes text translation training. }
    \label{fig:model}
    \vspace{-1em}
\end{figure*}

\textbf{Models.}
We apply multimodal translation design on two types of Speech-LLM architectures, SALM-T5 and BESTOW-GPT. 
SALM model~\cite{chen2024salm} is an open-source speech-language model proficient in speech recognition, word boosting, and translations. This model is designed to leverage a pre-trained and instruction-fine-tuned LLM by conditioning it on paired speech and text prompts to generate textual outputs for various speech tasks. We follow this architecture but replace the backbone LLM to a T5 architecture based machine translation pretrained model.  The resultant architecture is shown in Figure~\ref{fig:model}(a) and denoted as SALM-T5.
BESTOW is a cross-attention variant of SALM proposed in~\cite{chen2024bestow}. Unlike SALM, which prepends speech prompts to text prompts as LLM inputs, BESTOW applies cross-attention between speech prompts and the original textual LLM inputs at every step before feeding them into the LLM. In this cross-attention mechanism, the query consists of the original LLM inputs, and the key and value are the speech prompts, similar to the LAS model design for end-to-end ASR~\cite{chan2016listen}. We follow~\cite{chen2024bestow} and use TinyLlama as the LLM backbone. The resultant architecture is shown in Figure~\ref{fig:model}(b), denoted as BESTOW-GPT.
We introduce text translation task to the Speech-LLM training by bypassing the speech encoder and cross-attention transformer and only backpropagating the LLM backbone, as demonstrated in the figure. 

\textbf{Stationary data blending with a stochastic weighted multiplexer.}
The main objective for data blending is to ensure a stationary distribution of data sources in mini-batches throughout each training step. This ensures there are no local domain shifts throughout training, effectively stabilizing training and keeping model’s performance balanced the training domains, languages, and tasks.
With large training data random access sampling and blending during preprocessing are not viable; for the sake of brevity, we refer the reader to related discussion in~\cite{post-etal-2023-sotastream}.
Instead, we treat each individual dataset as a single stream read sequentially with some weight proportional to sampling likelihood. 
To blend data, we interleave the input streams into a single output stream using a stochastic weighted multiplexer (MUX).
At each iteration step, MUX randomly selects an input stream to yield example from according to a multinomial distribution defined by the input stream weights (similar to ~\cite{post-etal-2023-sotastream}). 
Note that with MUX, all input streams (datasets) are distributed uniformly throughout the training (i.e., stationary data source distribution).
We start with natural weights that are proportional to dataset size.
To allow upsampling and downsampling, we make each input stream infinite: once exhausted we iterate it again, re-shuffling underlying dataset shards for better randomization. 

\textbf{Sampling stratification on sequence length.}
To maximize the training efficiency, we use dynamic bucketing from Lhotse~\cite{zelasko2021lhotse}. It is a sampling strategy that groups utterances of similar sequence length in the same mini-batches, minimizing the necessary padding\footnote{
We acknowledge that stratification by sequence length may degrade the stationarity of data source distribution discussed in an earlier section, as datasets may have different sequence length distributions, leading to non-uniform data source representation in some buckets.
}. For GPT-based models, source and target text is usually collapsed together into a single sequence with context and answer, making it possible to use regular bucketing effectively. However, with encoder-decoder text and multimodal models, source and target lengths need to be considered separately. We found that stratifying only on source length leads to about 40\% padding on decoder's sequence length dimension. To address that issue, we propose a 2D bucketing strategy: we allocate a training example to a bucket based on the input sequence length, and then choose a sub-bucket based on the output sequence length. The 2D bucket bins are estimated based on a 500k sample of the data so that each bucket (and sub-bucket) contains approximately the same total number of seconds or tokens. This helps sample from buckets more uniformly since longer sequence buckets have smaller batch sizes. 

\textbf{Batch size OOMptimizer.}
With bucketing, each bucket requires a different batch size (inversely proportional to the sequence length) to use hardware efficiently. With 2D bucketing and 10x10 buckets, batch size calibration becomes tedious and inaccurate. We propose an automatic search algorithm to determine an optimal batch size profile for model, data, and hardware combination. The algorithm is a variant of binary search: we start with an initial batch size setting and run a full training step of the model (forward pass, backward pass, and parameter update) with random inputs to get an accurate estimate of the GPU memory utilization during training. If an out of memory (OOM) error is raised, we mark the batch size as invalid and halve it. If the training step succeeds, we mark the batch size as valid and double it. Once we have determined at least one of each (valid and invalid) batch sizes, we explore the middle point settings recursively until convergence. We terminate the search when the found valid batch size is within 5\% of invalid batch size.

\textbf{Joint multimodal training: combining modalities with round robin and zip samplers.}
The previous paragraphs described the process of sampling a stratified mini-batch for a single modality. 
We propose to sample a mini-batch for each modality separately, and combine them. We found two approaches to be effective: a randomized round robin sampler, which randomly chooses a singe modality at each training step; and a zip sampler, which combines two single-modal mini-batches into one, resulting in multimodal gradient accumulation. Our preliminary experiments demonstrated only minor differences, likely due to momentum-based optimizer enabling both modalities to participate in parameter updates for round robin approach. All multimodal experiments reported in this work use the round robin approach.

\section{Experimental setup}

At a high level, we demonstrate the effectiveness of our methods as follows. First, we score the baseline NMT and AST models on standard benchmarks. Then, we extend the NMT baselines to audio modality either through audio-only or multimodal training, and check how well they score on both modalities after the procedure. We use 128 NVIDIA A100 80GB GPUs for each experiment.

\textbf{Speech data}. We follow~\cite{canary1b} to include 31k hours of public data speech and 54k hours of extra in-house data for speech recognition (ASR), including English (67.4k hours), German (6.1k hours), Spanish (6.6k hours), and French (5.1k hours).
We obtain AST data by generating synthetic labels for ASR data using NMT models~\cite{MegatronNMTEnAny2024, MegatronNMTAnyEn2024} in all 4 languages. 
Further, a 4.8k hours pseudo-labeled English $\rightarrow$ German translation dataset from~\cite{mittal2023leveragingsynthetictargetsmachine} was also used. 

\textbf{Text translation data}. The training set comprises public datasets\footnote{CC Aligned, CC Matrix, Paracrawl, WikiMatrix, EuroPat, Europarl, WMT CommonCrawl, UNPC, News Commentary, EESC, Rapid, EMEA, Covost, WikiTitles, and NLLB.} and synthetic data. For synthetic data generation, we performed back-translation with~\cite{MegatronNMTEnAny2024, MegatronNMTAnyEn2024} for all language pairs using monolingual data from CulturaX~\cite{nguyen2023culturaxcleanedenormousmultilingual}. To ensure accurate autoregressive modeling, we used synthetic translations as inputs and real text as targets.
For English-centric directions, the synthetic data make up 39\% of the total. Synthetic datasets are fully utilized for directions where a large parallel corpus is unavailable. For preprocessing, we first performed language identification. We then filtered out data entries that either fell outside the length range of 1 to 256 tokens or where the target sentence was longer than 1.3 times the length of the source sentence. Next, we applied Bicleaner~\cite{prompsit:2020:EAMT} to remove sentence pairs with a translation likelihood score below 0.5. These preprocessing steps resulted in a training set of 2.7 TB across 33 languages.

\textbf{Speech-LLM setup}.
For experiments we use PyTorch and NVIDIA NeMo~\cite{kuchaiev2019nemo}. The speech encoders of SALM-T5 and BESTOW-GPT are initialized from the Canary-1B model~\cite{canary1b}. SALM-T5 initializes the LLM backbone from T5 NMT model below and directly feeds pretrained speech encoder outputs to LLM without modality adapter layers. 
Following~\cite{chen2024bestow}, BESTOW-GPT initializes from TinyLlama-1.1B-chat model~\cite{zhang2024tinyllamaopensourcesmalllanguage} and uses a 2-layer cross-attention Transformer module.
We train all parameters in both models, which are both about 1.8 billion. We use distributed fused Adam~\cite{kingma2014adam} optimizer and cosine annealing, with learning rate $1e-4$ and weight decay of $1e-3$. Gradient clipping of $1.0$ is  applied. The data sampler uses 10x10 2D bucketing configuration, natural weights for data blending, and round robin modality combination with equal selection probabilities. %

\textbf{NMT LLM setup}. For our T5 NMT baseline, we adopted a multilingual NVIDIA Riva Megatron NMT model~\cite{rivamegatron} trained for general purpose translation. The text translation model follows the T5 architecture with the following modifications. It comprises 12 encoder layers and 24 decoder layers with 20 attention heads. The hidden dimension is 1280, and the feedforward layer has a dimension of 5120. We use a SentencePiece tokenizer with a vocabulary size of 64k, shared across the source and target languages. %
RMSNorm~\cite{zhang2019rootmeansquarelayer} is employed for normalization across layers. For optimization we leveraged fused Adam and an inverse Square Root Annealing learning rate schedule, starting with an initial learning rate of $4e-4$. Gradient clipping was applied at a threshold of 1.0 to stabilize training. Warmup steps were configured to 0.8\% of 3.6 million maximum steps. 

\textbf{Evaluation}. 
We evaluated AST performance on the public benchmarks FLEURS~\cite{conneau2023fleurs}.
The text translation NMT model is evaluated on FLORES-101~\cite{goyal2022flores} benchmark. 
We limit the evaluation to 6 pairs in 4 languages present in our AST data. 
Decoding is performed using greedy-search, and we report BLEU score using SacreBLEU tool~\cite{post2018call}. 

\section{Results}

\begin{table}
\caption{Speech translation SacreBLEU scores on a subset of FLEURS. The baselines are SeamlessM4T-v2~\cite{barrault2023seamless}, Canary-1B AST~\cite{canary1b}, and Canary-1B ASR cascaded with pretrained T5 NMT. BESTOW-GPT and SALM-T5 are TinyLlama and T5 models finetuned exclusively on speech. Multimodal indicates joint finetuning on speech and text. \label{tab:ast_fleurs}}
\begin{tabular}{|l|lll|lll|l|}
\hline
\multirow{2}{*}{Model}    & \multicolumn{3}{c|}{En $\rightarrow$} & De    & Es    & Fr                      & \multirow{2}{*}{Avg}  \\
                          & De    & Es    & Fr      & \multicolumn{3}{c|}{$\rightarrow$ En}                 &      \\ \hline%
\textit{Baselines}        &       &       &         &                         &       &       &      \\
SeamlessM4T-v2            & 33.2  & 23.7  & 43.0    & 37.1                    & 25.4  & 30.9  & 32.2 \\
Canary-1B                 & 32.1  & 22.7  & 40.8    & 34.0                    & 21.8  & 31.0  & 30.4 \\
\textit{\quad+T5 cascade} & 32.6  & 23.9  & 42.6    & \textbf{39.4}                    & 27.3  & \textbf{37.9}  & 34.0 \\ \hline%
BESTOW-GPT                & 32.0  & 23.1  & 41.2    & 35.8                    & 23.9  & 35.1  & 32.0 \\
\textit{\quad+multimodal} & 33.2  & \textbf{24.5}  & 42.7    & 37.9                    & 24.9  & 36.8  & 33.3 \\ \hline%
SALM-T5                   & 34.2  & 24.1  & 43.5    & 38.8                    & 25.9  & 37.0  & 33.9 \\
\textit{\quad+multimodal} & \textbf{34.6}  & \textbf{24.5}  & \textbf{44.6}    & 37.9                    & \textbf{27.5}  & 37.4  & \textbf{34.4} \\ \hline
\end{tabular}
\end{table}

\begin{table}
\caption{Text translation SacreBLEU scores on a subset of FLORES. The T5 NMT baseline is trained exclusively on text, and extended to audio modality as SALM-T5.  \label{tab:nmt_flores}}
\begin{tabular}{|l|lll|lll|l|}
\hline
\multirow{2}{*}{Model}     & \multicolumn{3}{c|}{En $\rightarrow$} & De    & Es    & Fr                      & \multirow{2}{*}{Avg}  \\
                           & De    & Es    & Fr      & \multicolumn{3}{c|}{$\rightarrow$ En}                 &      \\ \hline%
T5 NMT                     & 38.1  & 27.5  & 50.2    & 44.7                    & 30.3  & 45.7  & 39.4 \\
SALM-T5                    & 0     & 0     & 0       & 0                       & 0     & 0     & 0    \\
\textit{\quad+multimodal}  & 38.7  & 27.5  & 50.4    & 44.1                    & 30.9  & 45.8  & 39.6 \\ \hline
\end{tabular}
\end{table}

\textbf{Speech translation (Table~\ref{tab:ast_fleurs}).} Both BESTOW-GPT and SALM-T5 are trained purely for speech translation as baselines. We expect that these models "catastrophically forget" their text translation skills while they learn speech translation. With speech-only training, SALM-T5 naturally achieves a better mean BLEU score of 33.9 compared to BESTOW-GPT's 32.0 since the underlying foundation model was trained specifically for NMT. 
We confront these results with multimodal training, where we observe 1.3 absolute BLEU score improvement for BESTOW-GPT and 0.5 absolute BLEU score improvement for SALM-T5. Larger improvement for BESTOW-GPT makes sense since TinyLlama model hasn't had as much prior training on NMT data. In both cases, speech translation capability is improved by joint multimodal training, demonstrating the synergy between two modalities.

\textbf{Text machine translation (Table~\ref{tab:nmt_flores}).} We compare the performance of the T5 NMT model in three scenarios: pretrained text-only model, after speech-only finetuning (SALM-T5) and after joint multimodal training (SALM-T5+multimodal). First, we notice that SALM-T5 completely lost its capability of translating text while acquiring AST skills (as seen in Table~\ref{tab:ast_fleurs}). For multimodal SALM-T5, the mean BLEU score across 6 language pairs is preserved (we don't consider 0.2 as a meaningful improvement), while we observe up to 0.6 BLEU score variation in some pairs. Therefore, the first goal of multimodal training is achieved: the T5 model did not lose its original performance on text after extension to audio modality.

\begin{table}
\caption{Multimodal training efficiency gains from 2D bucketing and OOMptimizer (opt).  \label{tab:perf}}
\begin{tabular}{|l|l|l|ll|l|}
\hline
\multirow{2}{*}{Model} & \multirow{2}{*}{Steps} & \multirow{2}{*}{Runtime} & \multicolumn{2}{c|}{Mean batch size} & \multirow{2}{*}{Steps/sec} \\
 & & & Audio & Text &  \\\hline
BESTOW-GPT & 450k & 7 days & 13 & 100 & 1.15 \\  %
\textit{\quad+opt} & 100k & 2.5 days & 55 & 269 & 0.81 \\  %
SALM-T5\textit{+opt} & 7.5k & 5 hours & 30 & 251 & 0.7 \\\hline  %
\end{tabular}
\end{table}

\textbf{Efficient multimodal training with 2D bucketing and OOMptimizer.} 
We compare the training efficiency of BESTOW-GPT without and with our optimizations in Table~\ref{tab:perf}. 
The baseline used 1D bucketing with a batch size heuristic on the total number of seconds (audio) or tokens (text) in a mini-batch, and a quadratic length penalty to offset the quadratic memory complexity of transformer. 
The penalty and total length thresholds were tuned manually to observe close to 100\% allocated GPU memory. 
SALM-T5 model was trained directly with OOMptimizer to save resources and converged in 5 hours, which is much faster than BESTOW-GPT likely due to being pretrained specifically for NMT. We noticed that without OOMptimizer, the bucket batch sizes are constrained either by sequence length outliers (especially for drastically different input and output sequence length examples) or the longest sequence lengths in the dataset. 
2D bucketing increased the batch sizes for most buckets by 1.5-2x for audio modality and 2-4x for text modality when training SALM-T5 model, since the large sequence length outliers have their own sub-buckets with a smaller batch size. With BESTOW-GPT, we applied 2D bucketing for audio modality with a modest 15-20\% batch size increase.

\section{Conclusions}

We introduced EMMeTT, an efficiency-driven mutlimodal training regime. We extended two foundation models: Megatron-NMT-1B based on T5 (encoder-decoder) and TinyLlama GPT (decoder-only) to support speech translation. Text translation capabilities were kept intact with MUX data blending and a round robin modality selection strategy. 2D bucketing and batch size optimization substantially accelerated the training in the presence of variable sequence lengths. Although not demonstrated in this work, we expect this framework to perform effectively with larger model sizes and model parallelism techniques such as tensor/sequence/pipeline parallel or fully-sharded data parallel. We aim to extend EMMeTT to general-purpose multimodal LLM training in our future works. Model training code, together with OOMptimizer, is open-source and available as a part of NVIDIA NeMo toolkit.

\bibliographystyle{IEEEtran}
\bibliography{main}

\end{document}